%% file: main.tex
\documentclass[sigconf]{acmart}

\AtBeginDocument{%
  }

\setcopyright{acmlicensed}
\copyrightyear{2024}
\acmYear{2024}
\acmDOI{10.1145/3664647.3680595}

\setcopyright{acmlicensed}\acmConference[MM '24]{Proceedings of the 32nd ACM International Conference on Multimedia}{October 28-November 1, 2024}{Melbourne, VIC, Australia}
\acmBooktitle{Proceedings of the 32nd ACM International Conference on Multimedia (MM '24), October 28-November 1, 2024, Melbourne, VIC, Australia}
\acmISBN{979-8-4007-0686-8/24/10}

\usepackage{amsmath}
\usepackage{mathtools}
\usepackage{amsthm}
\usepackage{multirow}
\usepackage{pifont}       
\usepackage{bbding}
\usepackage{bm}
\usepackage{fontawesome}  
\usepackage{graphicx}
\usepackage{subcaption}
\begin{document}

\title{Student-Oriented Teacher Knowledge Refinement for \\ Knowledge Distillation}

\author{Chaomin Shen}
\orcid{0000-0001-9389-6472}
\affiliation{%
  \institution{School of Computer Science and Technology \\
  East China Normal University}
  \city{Shanghai}
  \country{China}
}
\email{cmshen@cs.ecnu.edu.cn}

\author{Yaomin Huang}
\orcid{0000-0001-9389-6472}
\affiliation{%
  \institution{School of Computer Science and Technology \\
  East China Normal University}
  \city{Shanghai}
  \country{China}
}
\email{ymhuang@stu.ecnu.edu.cn}

\author{Haokun Zhu}
\orcid{0009-0000-3558-0402}
\affiliation{%
  \institution{School of Computer Science and Technology \\
  East China Normal University}
  \city{Shanghai}
  \country{China}
}
\email{52205901005@stu.ecnu.edu.cn}

\author{Jinsong Fan}
\orcid{0009-0009-0851-8924 }
\affiliation{%
  \institution{Wenzhou University}
  \city{Wenzhou}
  \country{China}
}
\email{fjs@wzu.edu.cn}

\author{Guixu Zhang}
\authornote{Corresponding author.}
\orcid{0000-0003-2568-4691}
\authornotemark[0]
\affiliation{%
  \institution{School of Computer Science and Technology \\
  East China Normal University}
  \city{Shanghai}
  \country{China}
}
\email{gxzhang@cs.ecnu.edu.cn}

\renewcommand{\shortauthors}{Chaomin Shen, Yaomin Huang, Haokun Zhu, Jinsong Fan, \& Guixu Zhang}

\begin{abstract}
Knowledge distillation has become widely recognized for its ability to transfer knowledge from a large teacher network to a compact and more streamlined student network.
Traditional knowledge distillation methods primarily follow a teacher-oriented paradigm that imposes the task of learning the teacher's complex knowledge onto the student network. However, significant disparities in model capacity and architectural design hinder the student's comprehension of the complex knowledge imparted by the teacher, resulting in sub-optimal performance.
This paper introduces a novel perspective emphasizing student-oriented and refining the teacher's knowledge to better align with the student's needs, thereby improving knowledge transfer effectiveness.
Specifically, we present the Student-Oriented Knowledge Distillation (SoKD), which incorporates a learnable feature augmentation strategy during training to refine the teacher's knowledge of the student dynamically.
Furthermore, we deploy the Distinctive Area Detection Module (DAM) to identify areas of mutual interest between the teacher and student, concentrating knowledge transfer within these critical areas to avoid transferring irrelevant information. This customized module ensures a more focused and effective knowledge distillation process.
Our approach, functioning as a plug-in, could be integrated with various knowledge distillation methods. Extensive experimental results demonstrate the efficacy and generalizability of our method.
\end{abstract}





\maketitle
\begin{figure}[t]
\centering
\includegraphics[width=\columnwidth]{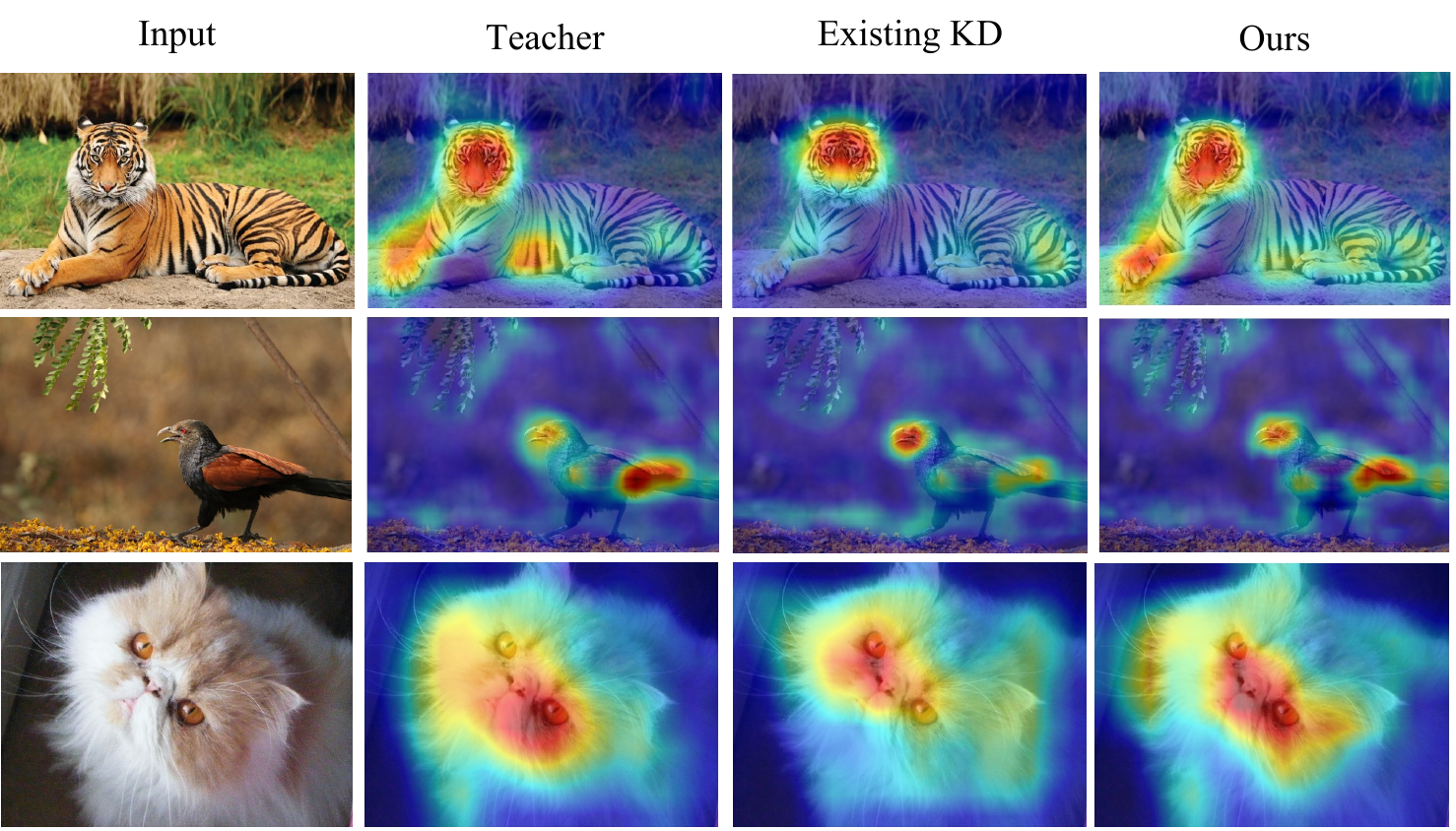}
\caption{
This research tackles the issue where a powerful teacher network identifies key data regions, but its smaller student struggles to understand these patterns. Our method refines the teacher's knowledge for the student, helping it better recognize these patterns.
}
\label{motivation}
\end{figure}
\section{Introduction}
\label{Intro}
Knowledge distillation, first introduced by~\cite{hinton2015distilling}, has attracted significant interest in both academic and industrial research for its effectiveness in transferring knowledge from a pre-trained
and high-performance teacher network into a more compact and lower-capacity student network. This knowledge transfer improves the student network's performance while preserving its structure.
Knowledge distillation has been applied in various tasks such as classification~\cite{li2023curriculum, jin2023multi}, object detection~\cite{chen2017learning, yang2022focal}, and semantic segmentation~\cite{liu2019structured, shu2021channel}. 

The original knowledge distillation~\cite{hinton2015distilling} leverages soft labels provided by the teacher network to guide the student network.
Subsequently, logits-based knowledge distillation has investigated various constraints through decoupled logits~\cite{zhao2022decoupled, yang2023knowledgewsld} and more comprehensive constraints~\cite{jin2023multi, gou2023multi}. 
Since logits only provide information on the distribution at the class level and lack the comprehensive structural information of the input data, feature-based knowledge distillation~\cite{romero2014fitnets}, which distills the features through pixel-level constraints applied at the intermediate layers, has increasingly gained attention~\cite{zagoruyko2016paying, tian2019contrastive, chen2021distilling}.

Figure~\ref{motivation} utilizes Grad-CAM~\cite{selvaraju2017grad} to visualize the crucial regions prioritized by the network, enabling an assessment of recognition patterns across different networks.
The results indicate that due to substantial disparities in model capacity and architecture design, it is challenging for the student to fully understand the recognition patterns of the teacher from the intricate teacher knowledge, ultimately leading to sub-optimal performance.
Existing approaches often facilitate the student's understanding of complex knowledge from the teacher via surrogate representation~\cite{mirzadeh2020improved, zagoruyko2016paying, yim2017gift, tung2019similarity, lin2022knowledge}, or by implementing rigid constraint~\cite{yang2022masked, tian2019contrastive, heo2019comprehensive}. 
All these methods adopt a teacher-oriented perspective, assuming that the teacher's knowledge is fully applicable and beneficial to the students, neglecting the inherent differences in their capabilities and structural designs. 
Given this critical insight, we propose shifting to a student-oriented perspective that tailors teacher knowledge to the student's learning capabilities and architectural design.
The core issue in our method is: \textit{how to appropriately adjust the teacher's knowledge within a reasonable scope to adapt to the needs of the student network.}

Data augmentation, known for diversifying input data through various transformations—emerges as a promising approach.
Its ability to generate new data from the same distribution 
\cite{DBLP:journals/jmlr/BengioBBBBCCCEEGMLPRSS11}
allows for refining the teacher's knowledge to better match the student's needs while preserving the teacher's original knowledge to avoid loss.
However, since enhancements at the input level are not directly related to the distilled knowledge (e.g., features and logits), the impact of data augmentation on input data for distillation remains uncontrollable.
Furthermore, ~\cite{DBLP:conf/icml/BengioMDR13, DBLP:journals/corr/OzairB14} demonstrated that feature augmentation in high-dimensional spaces offers the advantage of increased plausibility of 
generated data points, thereby enhancing the likelihood of producing reasonable results.
Therefore, our strategy shifts towards leveraging the potential of augmentation at a finer level of granularity within the latent space. By enhancing the features of latent space, we aim to directly tailor the teacher's knowledge, making it more accessible and relevant for the student network.
Considering the manually selected augmentation strategies not only require a significant amount of grid search time to find the optimal strategy, but they may also disrupt the distribution of the original teacher knowledge and cannot guarantee that the augmented features will be suitable for the student network.
Inspired by neural network search~\cite{zoph2016neural}, automatically searching for the optimal augmentation strategy provides a great idea.
This automated search for feature-level augmentation strategies can avoid introducing human biases, prevent unreasonable augmentation strategies from undermining the original knowledge, and significantly reduce the time spent on grid searches for various augmentation strategies.

Based on the above-mentioned analysis, the core idea of our proposed method is: \textit{adjust the teacher's knowledge through a learnable feature augmentation strategy}.
Specifically, we introduce Student-Oriented Knowledge Distillation (SoKD), an innovative perspective that dynamically tailors the pre-trained teacher network's knowledge to the needs of the student network.
SoKD consists of two key components: Differentiable Automatic Feature Augmentation (DAFA) and
the Distinctive Area Detection Module (DAM).
DAFA is guided by student knowledge, searching for the most suitable augmentation strategy within a carefully designed feature augmentation search space.
DAM utilizes shared parameters to identify areas of mutual interest between the teacher and student, facilitating knowledge transfer and easing the student's learning process.
In summary, the main contributions of the paper are:
\begin{itemize}
\item From a student-oriented perspective, we proposed that SoKD adjusts teacher knowledge to accommodate the capacity and architectural design of the student network while preserving the overall integrity of the original teacher's knowledge.

\item We apply DAFA to automatically learn the most suitable enhancement strategy for adjusting the teacher's knowledge through an automated search method. We utilize DAM to identify mutual distillation areas, improving information transfer efficiency and simplifying the student's learning process.

\item SoKD can be plugged into existing knowledge distillation methods, and extensive experiments show that SoKD can significantly improve the performance of these methods. 
\end{itemize}

\begin{figure*}[t]
\centering
\includegraphics[width=0.99\textwidth]{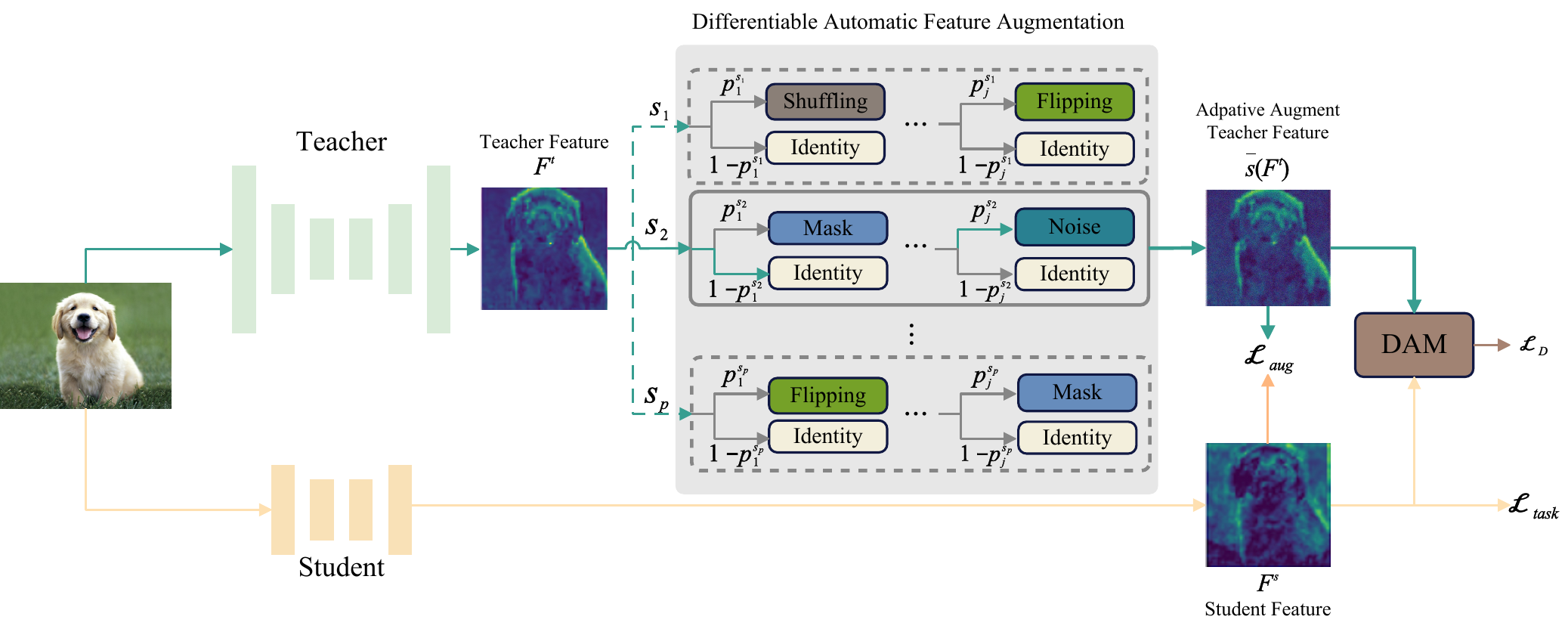}
\caption{
The overall framework of SoKD comprises two key components: 1) DAFA, a differentiable module for augmenting feature strategy search. This module adapts strategies during training, aiming to uncover knowledge more suitable for the student network. 2) DAM, which identifies distinctive areas between the teacher and student networks. This module focuses on areas of mutual interest for knowledge transfer, thereby avoiding unnecessary knowledge distillation.
}
\label{framework}
\end{figure*}

\section{Related Work}
\label{related_work}
\subsection{Knowledge Distillation}
Following the original work of knowledge distillation ~\cite{hinton2015distilling}, 
a series of studies \cite{zhang2018deep, mirzadeh2020improved, zhao2022decoupled, jin2023multi} improves the representation of logits. These logits-based methods transfer knowledge by minimizing the 
Kullback-Leibler divergence between 
the predicted logits of teachers and students.
The feature-based distillation methods \cite{romero2014fitnets} use features from intermediate layers. They have garnered more attention, as the higher-level logits-based methods lack structural information.
However, a substantial gap between the teacher and student prevents the latter from fully acquiring the comprehensive knowledge of the former.
Studies by ~\cite{gou2023multi, jin2023multi, li2023curriculum, chen2021distilling} promote the
student to learn knowledge as accurately as possible through more comprehensive and stringent constraints. In addition, many works ~\cite{wang2018progressive, mirzadeh2020improved, kim2021self} use a progressive distillation paradigm to 
avoid direct distillation when the gap between the teacher and student networks is large. Other methods ~\cite{tian2019contrastive, yang2022masked, srinivas2018knowledge} improve the transfer of knowledge to the student by refining the constraints.
While ~\cite{Liu_2020_CVPR, Dong_2023_CVPR} recognized that teacher knowledge might not suit the student, they searched student architectures adaptable to teacher knowledge from the student's perspective. However, searching for student network architectures is time-consuming and often yields architecture unfriendly to edge devices.

\subsection{Augmentation}
In the past few years, handcrafted data augmentation techniques have been widely used in training networks. For example, rotation, translation, cropping, resizing, and flipping are commonly used to augment training examples. Beyond these, techniques such as  Cutout~\cite{kim2021co}, Mixup~\cite{zhang2017mixup}, and CutMix~\cite{yun2019cutmix} are also adopted. 
Inspired by data augmentation~\cite{zhang2017mixup, yun2019cutmix}, current research boosts the network's representative ability by feature space augmentation. 
It is suggested
that higher-level representations amplify the volume of credible data points in the feature space
~\cite{bengio2013better, ozair2014deep}.
Given that features are usually well linearized~\cite{upchurch2017deep}, it is therefore feasible to use simple vector interpolation~\cite{gardner2015deep} and mixing up~\cite{verma2019manifold}.
Features are perturbed in the directions of intra-class/cross-domain variability ~\cite{li2021simple},
and instance features are directly synthesized by leveraging semantics ~\cite{chen2019multi}.
Although these methods achieve promising improvements on the corresponding tasks, they need expert knowledge to design the operations and set the hyper-parameters for specific datasets.
Recently, inspired by the neural architecture search (NAS)~\cite{zoph2016neural}, some methods attempted to learn data augmentation policies automately.
Note that ~\cite{liang2023exploring, yuan2024fakd} also use augmentation for knowledge distillation, 
their primary goal is to amplify the knowledge corresponding to the non-target categories in the label rather than achieving student-oriented knowledge adjustment. They expanded the teacher's knowledge, whereas this study seeks to tailor the teacher's knowledge to accommodate the students' requirements.

\section{Methodology}
\label{methods}
In this section, we will introduce our Student-Oriented Knowledge Distillation
(SoKD). Our method has two core components:
1) Differentiable Automatic Feature Augmentation (DAFA) in Section~\ref{sec.DFE}, and 2) Distinctive Area Detection Module (DAM)  in Section~\ref{sec.dam}.
The overall framework of SoKD is shown in Figure \ref{framework}.

\subsection{Differentiable Automatic Feature Augmentation}
\label{sec.DFE}
Our approach builds upon and improves the foundation of feature-based knowledge distillation.
For a given set of inputs $x$, the general form of the feature-based knowledge distillation is:
\begin{align}
\label{loss_feat_ori}
    \mathcal{L}_{\text{feat}} = (f^{t}(x) - g(f^{s}(x)))^2,
\end{align}
where $g(\cdot)$ is the mapping function transforming the student's feature to align with the teacher's feature, 
and $f^{t}$ and $f^{s}$ denote the teacher and student backbone blocks respectively. 
The total training objective for the student model is:
\begin{equation}
\label{loss_all_1}
    \mathcal{L}_{\text{train}} =
    \mathcal{L}_{\text{task}} + \alpha \mathcal{L}_{\text{feat}},
\end{equation}
where $\mathcal{L}_{\text{task}}$ is the standard task training loss for the student,
and $\alpha$ is the corresponding weight.

Given that the parameters in the pre-trained $f^{t}$ are fixed, the teacher network is limited to providing knowledge with its own bias. 
The student may have difficulties to understand 
this complex and fine-grained knowledge, and this kind of knowledge itself, may
often be inappropriate for the student. 
In this study, we aim to dynamically adjust the teacher network's knowledge $\mathcal{F}^t = f^{t}(x)$ to better suit the needs of the student network.

To preserve teacher network knowledge without changing parameters, feature-level augmentation is preferred over input augmentation, as higher-level representations expand the relative volume of plausible data points within the feature space~\cite{devries2017dataset}. To bypass the biases and time costs of manual enhancement, we introduce DAFA, a NAS framework that dynamically tailors augmentation strategies to student needs during distillation.
\paragraph{Feature Search Space}
We design a search space focused on feature representation, for simplicity and effectiveness. By analyzing existing state-of-the-art models, we develop a series of operations that can significantly enhance the robustness of feature representation, such as masking and adding noise.
To identify an enhancement strategy that meets the requirements within the minimum possible search time, we 
adopt a procedure inspired by Fast AutoAugment~\cite{lim2019fast}.
Given the knowledge $\mathcal F^t$ from the teacher, we wish to find a policy $s(\mathcal F^t)$ which could adaptively adjust the teacher's knowledge during the training process, thereby meeting the learning needs of the student at the current stage.

Suppose the policy $s(\mathcal F^t)$, denoted by $s$ for short, has $P$ sub-policies. 
Each sub-policy $s_i$, $1\le i \le P$, has
$k$ operations $O_j^{s_i}$ with the probability $p_j^{s_i}$ for $j=1,\cdots,k$, or do not do any operation, i.e., keep $\mathcal F^t$ unchanged. Combing these two cases, each $s_i$ corresponds to $k$ operations:    
\begin{equation}
\bar{O}_j^{s_i}\left(\mathcal{F} ; p_j^{s_i}, m_j^{s_i}\right)= 
\begin{cases}
    O_j^{s_i}\left(\mathcal{F} ; m_j^{s_i}\right)   &\text{with }  p_j^{s_i}, \\ 
    \mathcal{F}                             &\text{with }  1-p_j^{s_i},
\end{cases}
\end{equation}
for $j=1,\cdots,k$,
where $m_j^{s_i}$ is the magnitude of the operation. Thus, the 
complete sub-policy $s_i(\mathcal F)$ can be represented by:
\begin{equation}
s_i(\mathcal{F})=\bar{O}_k \circ \bar{O}_{k-1} \circ \cdots \circ \bar{O}_1(\mathcal{F}),
\end{equation}
for $i=1,\cdots, P$.

\input{table/homo_cifar}
\paragraph{Feature Search Strategy}
After describing the operations in the feature search space, we now focus on
the feature search strategy.
Given that the selection of sub-policies is a discrete process, to facilitate the end-to-end training, we should make the search space continuous.
Specifically, the sub-policy selection and operations are sampled from Categorical and Bernoulli distributions, respectively.
Therefore, we can directly determine the Top-k $s$ by predicting category probability and determining whether each operation $O$ is executed through the Bernoulli distribution.
To select a specific sub-policy $s$
and make the search space continuous, we relax the categorical choice of a particular operation to a softmax:
\begin{align}
    \bar{s}(\mathcal{F})=\sum_{s \in S} \frac{\exp \left(\alpha_s\right)}{\sum_{s^{\prime} \in S} \exp \left(\alpha_{s^{\prime}}\right)} s(\mathcal{F})
\label{eq_sp}
\end{align}
over all possible operations,
where $\mathcal{S}$ is the set of all candidate sub-policies, and 
$\bm\alpha=(\alpha_1,\cdots,\alpha_{|\mathcal{S}|})$ is a vector. At the end of the search, a discrete feature augmentation strategy can be obtained with the most likely operation, i.e., $s = \arg\max\limits_{s\in \mathcal{S}} \alpha_s$.

Consequently, searching for feature augmentation is simplified to learning a set of variables $\bm\alpha$ whose components are continuous.

After selecting a specific sub-policy using the above step, 
within this sub-policy, we determine whether this operation  is executed 
by sampling from a Bernoulli distribution. 
Essentially, this introduces a stochastic process, assigning a probability of execution or non-execution to each operation.
The feature operation $\bar{O}$ with the 
application probability $\beta$ and magnitude $m$ can be represented as:
\begin{equation}
s(\mathcal{F})=b\cdot O(\mathcal{F} ; m)+(1-b)\cdot \mathcal{F}, \quad b \sim \operatorname{Bernoulli}(\beta).
\end{equation}

After the relaxation process, the next step is to jointly optimize the feature augmentation strategy parameters $\gamma=\{\bm\alpha,\bm\beta, m\}$ and the student network weights $w$. 
We define $\mathcal{L}_{\text{train}}(w, \gamma)$ and $\mathcal{L}_{\text{aug}}(w,\gamma)$ as the training and validation losses, respectively. The aim is to find $\gamma^*$ that minimizes the augmentation loss $\mathcal{L}_{\text{aug}}$, with the optimal weights $w^*$ being derived by minimizing the training loss $w^*=\arg\min\limits_w \mathcal{L}_{\text {train }}(w, \gamma^*)$:
%
\begin{equation}
\begin{array}{ll}
\label{fss_loss}
\min\limits_{\gamma} \ \mathcal{L}_{\text {aug}}\left(w^*(\gamma), \gamma\right) \\
\text { s.t. }   w^*(\gamma)=
\arg\min\limits_w \mathcal{L}_{\text {train }}(w,\gamma).
\end{array}
\end{equation}
To estimate the gradient of $\mathcal{L}_{\text {aug}}$ with respect to 
parameters ${\bm\alpha, \bm\beta, m}$, the Gumbel-Softmax reparameterization trick is utilized to reparameterize the parameters ${\bm\alpha, \bm\beta}$, making the gradient differentiable. With the Gumbel-Softmax reparameterization,  Eq.~(\ref{eq_sp}) could be represented as:
\begin{align}
    \bar{s}(\mathcal{F})=\sum_{s \in S} \frac{\exp \left((\log(\alpha_s)+g_s)/\tau\right)}{\sum_{s^{\prime} \in S} \exp \left((\log(\alpha_{s^{\prime}})+g_{s^\prime})/\tau\right)} s(\mathcal{F}),
\label{eq_sp_gumbel}
\end{align}
where $g = -\log(-\log(u))$ with $u \sim \operatorname{Uniform}(0,1)$, and $\tau$ is the temperature of Softmax function.

Similarly, 
we apply the same reparameterization trick to the Bernoulli distribution: 
\begin{equation}
\begin{aligned}
\operatorname{Dis}(\lambda, \beta) &= \sigma\left(\left(\log \frac{\beta}{1-\beta}+\log \frac{u}{1-u}\right) / \lambda\right), \\
u & \sim \operatorname{Uniform}(0,1),
\end{aligned}
\end{equation}
such that the sigmoid function $\sigma$ 
is differentiable with respect to $\beta$.

\begin{figure}[t]
    \centering
    \includegraphics[width=0.99\linewidth]{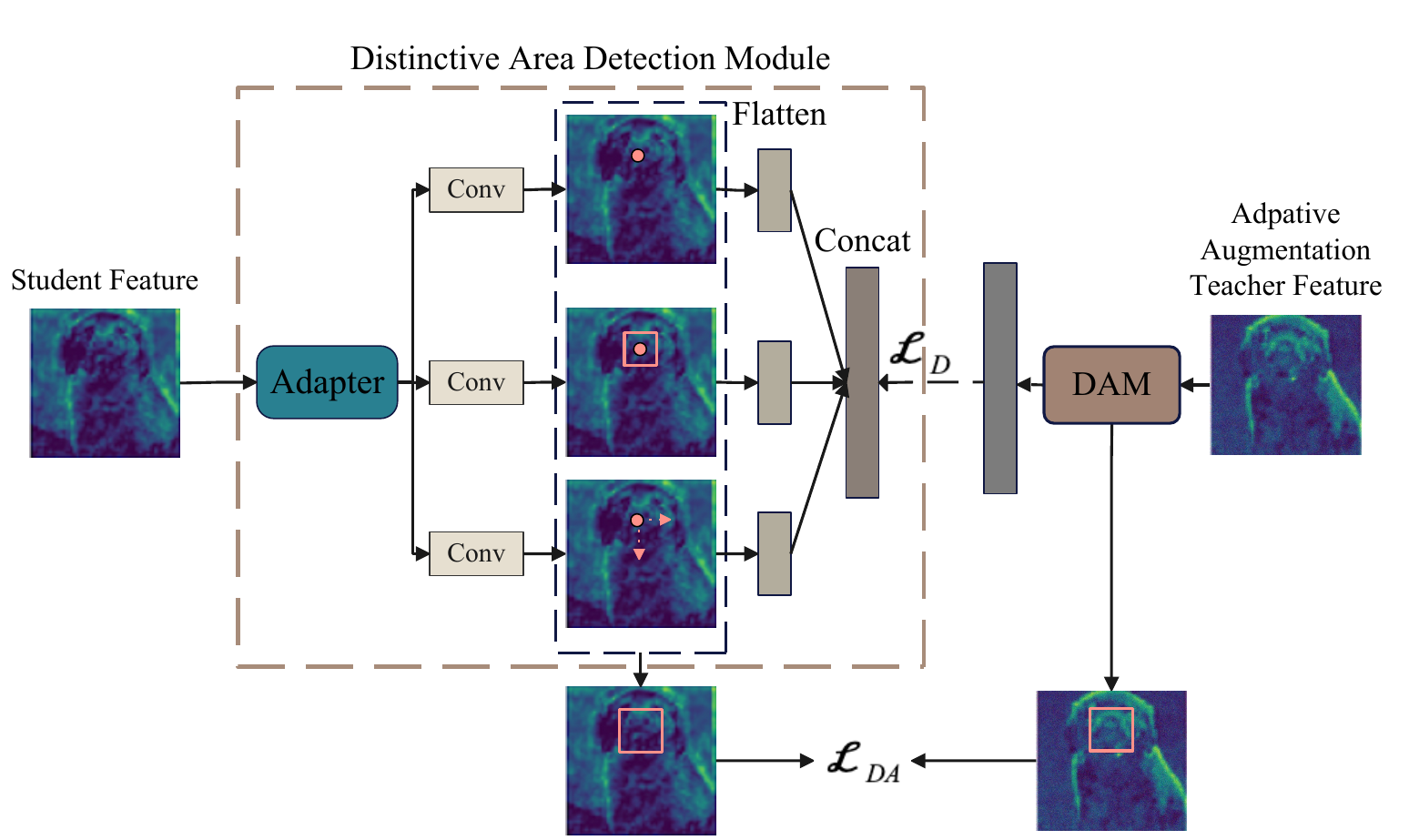}
    \caption{DAM in SoKD.
    Utilizing three head branches, DAM individually predicts the heatmap, size, and offset, thereby identifying the important areas of the feature. The teacher and student features are inputted into the corresponding DAM, which has an identical structure and shared parameters, aiming at identifying distinctive areas of mutual interest to both the teacher and student networks.}
    \label{DAM}
\end{figure}
Since some operations in the search space are non-differentiable, we employ a straight-through gradient estimator~\cite{bengio2013estimating} to optimize the augmentation magnitude $m$. 
For a feature $\hat {\mathcal{F}} = s(\mathcal{F})$ augmented by sub-policy $s$, the influence of the augmentation operation on each pixel $(i,j)$ of the image is uniform, specifically, $\frac{\partial \hat{\mathcal{F}}_{i, j}}{\partial m}=1$, the gradient of the magnitude can be calculated as:
\begin{equation}
\frac{\partial \mathcal{L}_{\text{aug}}}{\partial m}=\sum_{i, j} \frac{\partial \mathcal{L}_{\text{aug}}}{\partial \hat{\mathcal{F}}_{i, j}} \frac{\partial \hat{\mathcal{F}}_{i, j}}{\partial m}=
\sum_{i, j} \frac{\partial \mathcal{L}_{\text{aug}}}{\partial \hat{\mathcal{F}}_{i, j}}.
\end{equation}
Through the aforementioned reparameterization trick, we have transformed the non-differentiable feature search into a differentiable operation, hence making it possible to optimize parameters through gradient updates. 
\subsection{Distinctive Area Detection Module}
\label{sec.dam}
DAM is the second module of SoKD. As shown in the right part of Figure \ref{framework}, both the student's feature  and the teacher's feature after DAFA are fed into DAM, and DAM outputs the 
distinctive areas.
The purpose of DAM is to
address the challenge that
even when the bias in the knowledge provided by the teacher is alleviated by DAFA, fully replicating the teacher’s comprehensive information remains challenging.

\input{table/heme_cifar}
DAM aims to decouple the feature, thereby enabling the separation of distinctive areas for the transmission of knowledge in these distinctive areas.
Specifically, we first employ an adapter to facilitate the mapping of the student and teacher features into a joint semantic space.
Following this, we use a multi-branch detection head to pinpoint distinctive areas.
The DAM module comprises three branches, each consisting of consecutive convolutional layers, i.e., $3\times3$ and $1\times1$.
The output of DAM is:
\begin{equation}
A = \mathcal{D}(\operatorname{conv}_i(\Phi_s(\mathcal{F}^s))),
\end{equation}
where $\operatorname{conv}_i$ 
$(i=1,2,3)$ represents the three branches of the DAM module, $\Phi_s$ aligns the input between the teacher and the student, 
and $\mathcal{D}$ stands for the decode part of DAM,
which is used to generate distinctive areas.
During the training process, the features of both the student and teacher are passed through a shared-parameter DAM module to predict the distinctive areas independently. The predicted results are then supervised using $L_2$ loss, facilitating the alignment of distinctive areas between the student and teacher. 
The final training loss for the DAM module is:
\begin{equation}
\mathcal{L}_{D}=\left(\operatorname{conv}_i(\Phi_s(\mathcal{F}^s))-\operatorname{conv}_i(\bar{s}(\mathcal{F}^t))\right)^2.
\end{equation}

After  distinctive areas are filtered 
using DAM, Eq.~\ref{loss_feat_ori} is modified to the following representation:
\begin{equation}
\mathcal{L}_{\operatorname{DA}}\left(\mathcal{F}^s, \mathcal{F}^t\right)=\sum_{i=1}^{N}\left(\mathcal{M}(A_i)\Phi_s(\mathcal{F}^s)-\mathcal{M}(A_i)\bar{s}(\mathcal{F}^t)\right)^2,
\end{equation}
where $N$ is the number of the distinctive areas, and $\mathcal{M}$ is a mask operation that generates the corresponding mask based on the distinctive areas $A_i$.

\input{table/imagenet}
\subsection{Objectives for Optimization }
For $\mathcal{L}_\text{aug}$ in Eq.~(\ref{fss_loss}), 
to ensure that the enhanced feature provides suitable knowledge for the student, we use a consistency loss to make the features 
as close as possible to $\mathcal{F}^s$
after applying the corresponding sub-policy $s$. Therefore, the search strategy can adjust the teacher's knowledge to fit the student network:
\begin{align}
    \mathcal{L}_{\text{aug}} = \frac{1}{2}(s(f^t(x))-f^s(x))^2.
\end{align}

For the distillation of the student network, in addition to the loss related to the original task and the distillation loss at the feature level, we also carry out more coherent knowledge distillation based on DAM in Sec.\,\ref{sec.dam}. Thus, the final loss function can be expressed as:
\begin{equation}
\begin{array}{ll}
\underset{\gamma}{\min} 
& \mathcal{L}_{\text {aug}}\left(w^*(\gamma), \gamma\right)\\
\text { s.t. } & w^*(\gamma)={\arg\min\limits_w} \,(\mathcal{L}_{\text {task}}+ \alpha\mathcal{L}_{\text {D}} + \beta\mathcal{L}_{\text {DA}}),
\end{array}
\end{equation}
%
where $\alpha$, $\beta$ represent corresponding weights.

Through the optimization of the aforementioned bi-level problem, we can determine the optimal feature augmentation strategy, thereby optimizing the student network under the proposed distillation framework.

\section{Experiments}
\label{exp}
In this section, we first provide a detailed introduction to the implementations of our experiments. Subsequently, we conduct
comparisons with mainstream methods on various datasets and tasks. We also provide an analysis for further insights.

\subsection{Experimental Settings}
\paragraph{Baselines}
We conducted extensive comparative experiments on teacher-student pairs across various neural network architectures~\cite{simonyan2014very, he2016deep, zagoruyko2016wide, zhang2018shufflenet, ma2018shufflenet, howard2017mobilenets, sandler2018mobilenetv2} to validate the effectiveness of our method.

Our method can be integrated as a plug-in technique with various feature-based knowledge distillation approaches to enhance their performance. We applied our SoKD to existing distillation frameworks including FitNet~\cite{romero2014fitnets}, CRD~\cite{tian2019contrastive}, AT~\cite{zagoruyko2016paying} and ReviewKD~\cite{chen2021distilling}.

\paragraph{Datasets}
We employed three prominent datasets to evaluate our methodologies. The first dataset is the CIFAR-100 ~\cite{krizhevsky2009learning}, which includes 60,000 images in 100 unique classes. Each image is 32x32 pixels in resolution. The dataset is partitioned into two sections: a training subset with 50,000 images and a test subset with 10,000 images.
The second dataset is the ImageNet ~\cite{deng2009imagenet}, an essential dataset for image classification. It contains approximately 1.3 million training images and 50,000 validation images spread across 1,000 classes. The ImageNet dataset is notable for its high-resolution images.
Finally, the MS-COCO dataset ~\cite{lin2014microsoft} was also used for object detection tasks. This dataset includes images categorized into 80 classes, with a training set of 118,000 images and a validation set of 5,000 images.

\input{table/coco}
\paragraph{Implementation Details}
For CIFAR-100 experiments, we follow the basic settings of \cite{romero2014fitnets, zhao2022decoupled, chen2021distilling} with a batch size of 64, learning rate of 0.05, and SGD optimizer, training for 240 epochs on NVIDIA-A100. For ImageNet, we use a batch size of 512, an initial learning rate of 0.1, over 100 epochs, reducing the learning rate at epochs 30, 60, and 90. For COCO, we use the Detector2 framework \cite{wu2019detectron2} for object detection experiments. All codes are implemented based on PyTorch \cite{paszke2019pytorch}.

\subsection{Main Results}

\paragraph{CIFAR-100}
To fully demonstrate the efficacy of our method, we conducted extensive comparative experiments with various teacher-student pairs on CIFAR-100. Tables~\ref{homo_cifar} and \ref{heme_cifar} present the experimental results for homogeneous and heterogeneous architectures, respectively. 
The results indicate that SoKD significantly enhances the performance of original knowledge distillation in structurally similar teacher-student pairs (e.g., an improvement of 2.33 percentage points on FitNet for the ResNet56-ResNet20 pair). Moreover, it effectively transfers knowledge from the teacher to the student, even in pairs with larger structural differences (e.g., an improvement of 3.96
percentage points on FitNet for the ResNet50-MobileNetV2 pair).

\paragraph{ImageNet}
The results shown in Table~\ref{imagenet} illustrate that  SoKD can still perform satisfactorily on the more challenging dataset ImageNet. 
When employing ResNet34 as the teacher and ResNet18 as the student, our method improves the top-1 accuracy of AT from 70.69\% to 72.13\%. Notably, SoKD also significantly improves performance over the current state-of-the-art ReviewKD, increasing top-1 accuracy from 71.61\% to 72.53\%.

\paragraph{Object Detection}
We extend our experiments to object detection, another fundamental computer vision task. Using Faster-RCNN~\cite{ren2015faster}-FPN~\cite{lin2017feature} as the backbones and adopting average precision (AP),
AP50 as evaluation metrics, the results in Table~\ref{COCO} indicate a comprehensive enhancement of existing distillation methods through 
the combination with SoKD. This also effectively demonstrates the generalizability of our method.

\begin{figure*}[ht]
\centering
\includegraphics[width=0.8\textwidth]{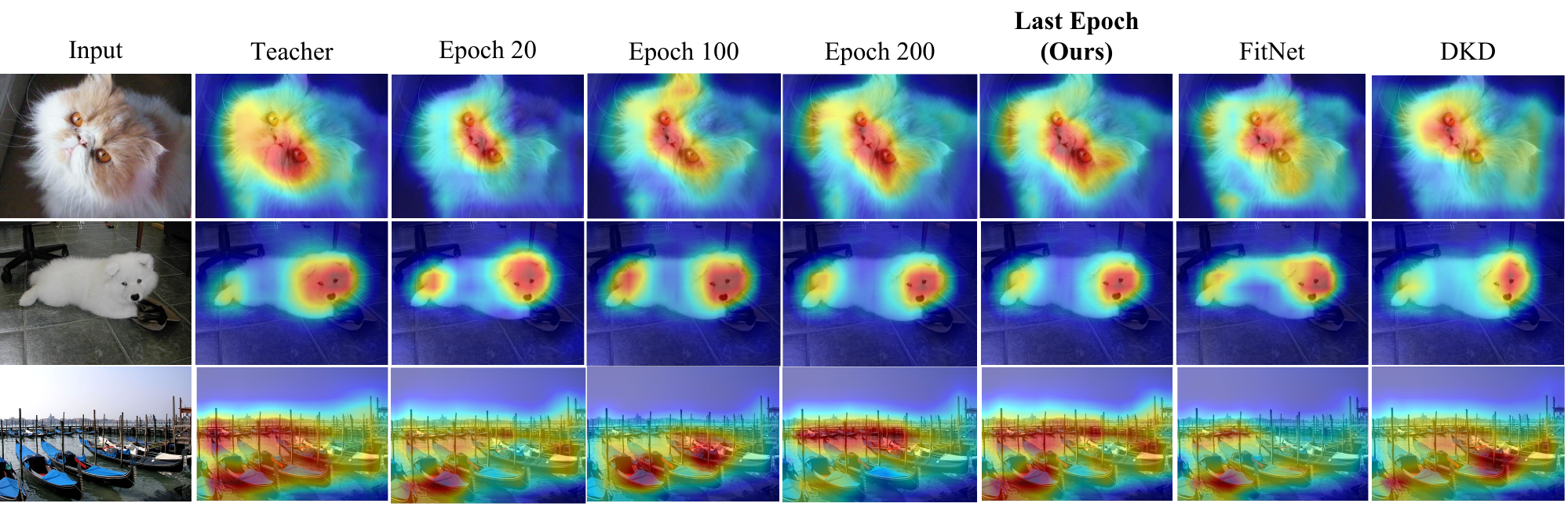}
\caption{In the distillation process on ImageNet, with ResNet34 serving as the teacher and ResNet18 as the student, the evolution of crucial regions within features. The final results are compared against FitNet and DKD.
}
\label{vis_grad}
\end{figure*}
\input{table/ablation_study}
\begin{figure}[t]
    \centering
    \includegraphics[width=0.8\linewidth]{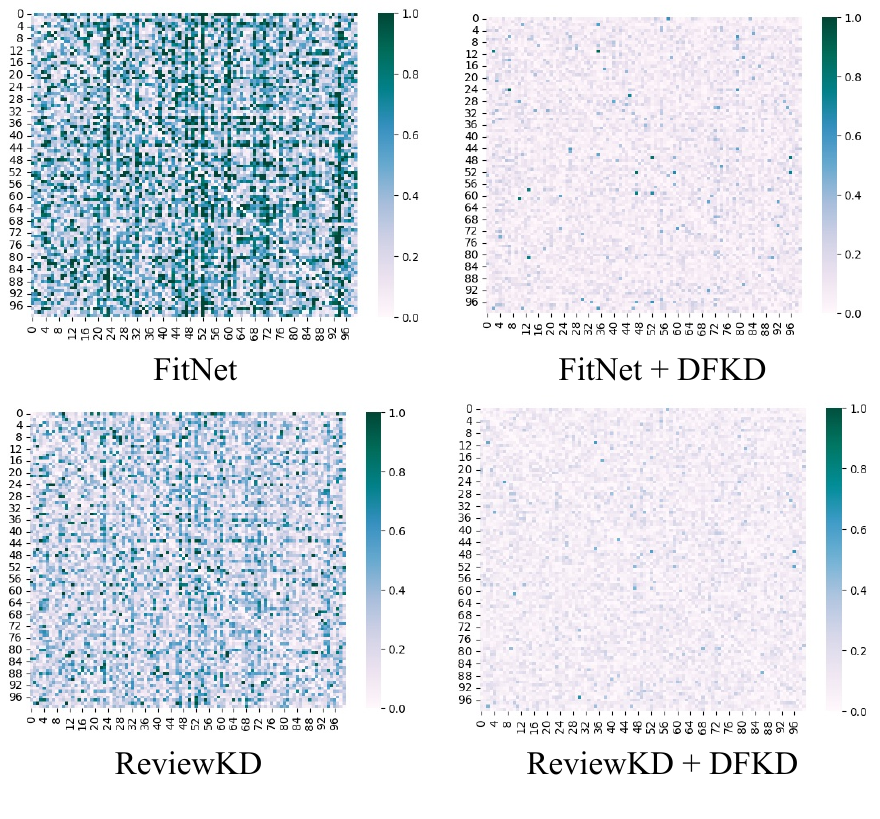}
    \caption{Difference of student and teacher outputs. SoKD leads to a significantly smaller difference than the baseline.}
    \label{cor_img}
\end{figure}

\subsection{More Analysis}
\paragraph{Ablation Study}
SoKD primarily comprises two crucial modules: DAFA 
and DAM.
Additionally, we enhance the training process of DAFA through a consistency loss $\mathcal{L}_{\text{aug}}$. We  validate the effect of each component using a one-by-one approach,
and conduct experiments with ResNet $32\times4$-ResNet $8\times4$ and VGG13-MobileNetV2, using ReviewKD as the baseline. The results in Table~\ref{main_ablation} demonstrate that each module within SoKD exhibits significant effectiveness.
\input{table/search_epochs}
\begin{figure}[t]
    \centering
    \begin{subfigure}[t]{0.45\linewidth}
        \centering
        \includegraphics[height=4cm,width=4cm]{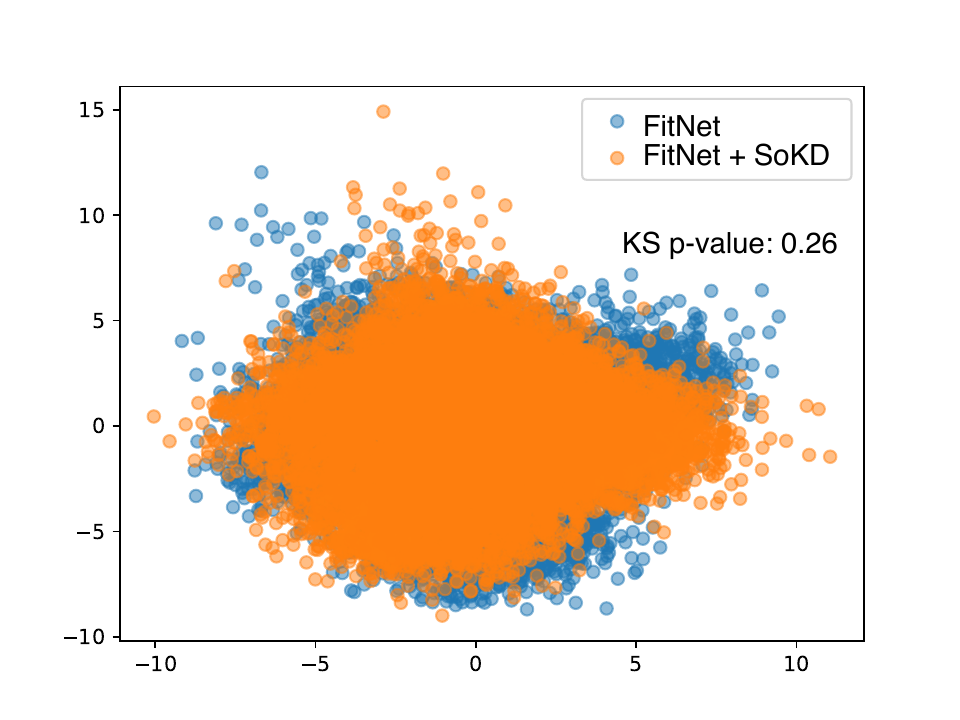}
        \caption{}
        \label{fig2.a}
    \end{subfigure}%
    \begin{subfigure}[t]{0.45\linewidth}
        \centering
        \includegraphics[height=4cm,width=4cm]{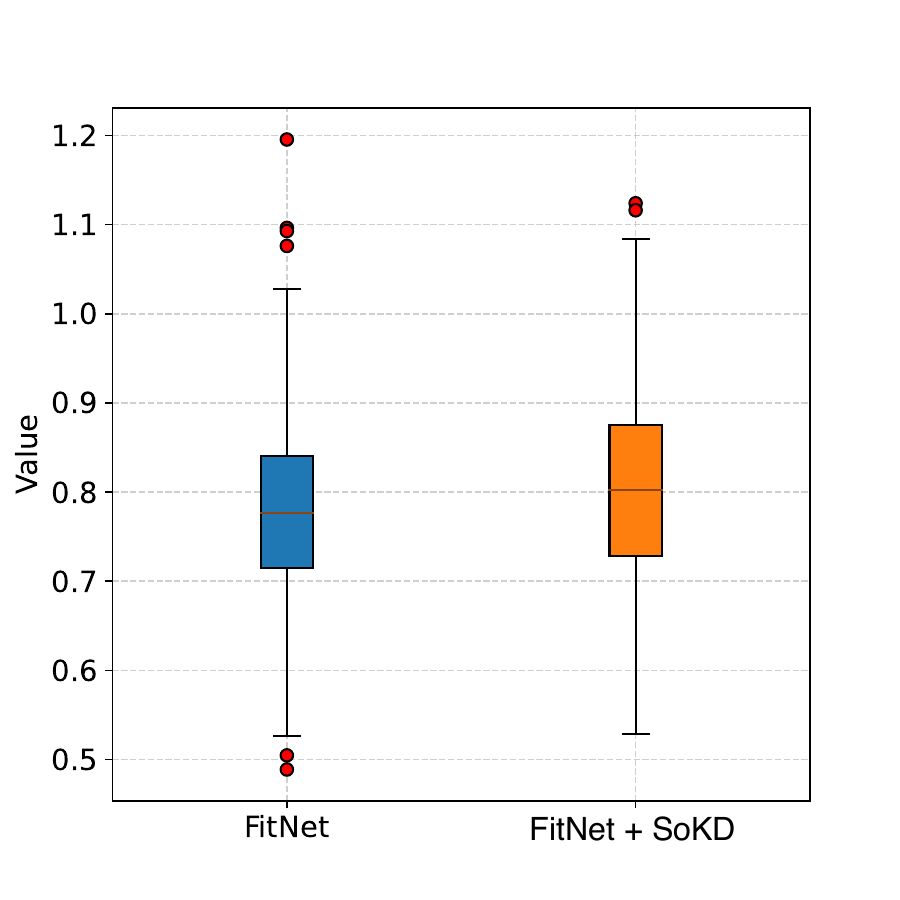}
        \caption{}
        \label{fig2.b}
    \end{subfigure}
    \caption{Visualization and statistical analysis of features. (a): Visualization of features after dimensionality reduction through PCA. (b): Box plot of features.
    The experiment was conducted on a teacher-student pair of ResNet $32\times4$ and ResNet $8\times4$ on the CIFAR100 dataset.}
    \label{feature_anaysis}
\end{figure}
\paragraph{Search epochs}
Excessive feature augmentation can undermine original knowledge, so we identify an optimal balance to enhance the teacher model's knowledge and guide the student model effectively. Table~\ref{search_epoch} shows comparative results for different search epochs on CIFAR-100 with ResNet110-ResNet32 and VGG13-MobileNetV2. Results indicate that only a few search epochs are needed to find the best feature augmentation strategy, as more epochs increase the risk of overfitting and are time-consuming.
\paragraph{Comparison of Manually Designed Augmentation Strategies}
\input{table/diff_aug}
To demonstrate our approach's superiority, we compared it with manual augmentation strategies, including direct input augmentation and simple feature augmentations (e.g., noise, masking, channel shuffling). Table~\ref{dif_aug} shows feature augmentation outperforms input augmentation, aligning with our expectations. However, simple augmentation alone does not yield significant improvements. Our method's core is adjusting the teacher's knowledge based on the student's needs, not just augmentation. The significant performance enhancement highlights our method's effectiveness.
\paragraph{Visualization}
Figure~\ref{vis_grad} uses Grad-CAM visualizations to show how the student network's focus areas change during the training process for ResNet34-ResNet18 distillation on ImageNet. The results show that the student gradually learns the teacher's recognition patterns. Compared to other methods, SoKD achieves a closer recognition pattern to the teacher.
Figure~\ref{cor_img} compares the final logits of teacher and student models. Previous methods, focusing on intermediate layers, often resulted in significant logit disparities. SoKD adjusts the teacher's intermediate knowledge, helping the student achieve similar logits, which is crucial for the final task.
\paragraph{Enhancing Pre- and Post-Feature Contrast}
Figure~\ref{fig2.a} visualizes the results before and after enhancing the features. The p-value of the Kolmogorov-Smirnov test is 0.26, which indicates that the overall distribution remains unchanged after the enhancement and no destruction of the original teacher's knowledge. 
The box plot results in Figure~\ref{fig2.b} suggest that the enhancement notably increased the diversity of features while preserving the original scope of knowledge and significantly reducing outlier occurrences, simplifying the student's task of capturing the teacher's knowledge and minimizing the risk of misleading information.
\section{Conclusion}
In this paper, we argue that the current teacher-oriented knowledge distillation often imposes the challenging task of learning complex teacher knowledge on the student network, frequently leading to sub-optimal outcomes. Therefore, we introduce a novel student-oriented knowledge distillation approach that employs automatically searched feature augmentation strategies. Without undermining the original knowledge of the teacher, this method appropriately adjusts the teacher's knowledge to accommodate the student network's model capacity and architectural design requirements. As a plug-in, SoKD significantly improves the performance of existing knowledge distillation methods on various datasets.


\bibliographystyle{ACM-Reference-Format}
\balance
\bibliography{main}

\end{document}

%% file: table/homo_cifar.tex
\begin{table*}[ht]
\centering
\caption{Results of top-1 accuracy (\%) for homogeneous architectures on CIFAR-100.}
\label{homo_cifar}
\begin{tabular}{ll|cccccc} 
\toprule
                         & Teacher  & ResNet56               & ResNet110              & ResNet$32\times4$      & WRN-40-2               & WRN-40-2               & VGG13                   \\
                         &          & 72.34                  & 74.31                  & 79.42                  & 75.61                  & 75.61                  & 74.64                   \\
                         & Student  & ResNet20               & ResNet32               & ResNet$8\times4$       & WRN-16-2               & WRN-40-1               & VGG8                    \\
                         &          & 69.06~                 & 71.14~                 & 72.50~                 & 73.26~                 & 71.98~                 & 70.36~                  \\ 
\cmidrule{1-2}\cmidrule{3-8}
\multirow{3}{*}{Logits}  & KD~\cite{hinton2015distilling}       & 70.66                  & 73.08                  & 73.33                  & 74.92                  & 73.54                  & 72.98                   \\
                         & DKD~\cite{zhao2022decoupled}      & 71.97                  & 74.11                  & 76.32                  & 76.24                  & 74.81                  & 74.68                   \\
                         & MLKD~\cite{jin2023multi}     & 72.19                  & 74.11                  & 77.08                  & 76.63                  & 75.35                  & 75.18                   \\
                         & ND$^*$~\cite{Sun2024Logit}     & 72.33                  & 74.32                  & 78.28                  & 76.95                  & 75.56                  & 75.22                   \\
\bottomrule
\multirow{8}{*}{Feature} & FitNet~\cite{romero2014fitnets}   & 69.21~                 & 71.06~                 & 73.50~                 & 73.58~                 & 72.24~                 & 71.02~                  \\
                         & +DFKD    & \textbf{71.54 (+2.33)} & \textbf{72.21 (+1.15)} & \textbf{74.41 (+0.91)} & \textbf{74.63 (+1.05)} & \textbf{73.02 (+0.78)} & \textbf{71.91 (+0.89)}  \\ 
\cmidrule{2-8}
                         & CRD~\cite{tian2019contrastive}      & 71.16~                 & 73.48~                 & 75.51~                 & 75.48~                 & 74.14~                 & 73.94~                  \\
                         & +DFKD    & \textbf{71.73 (+0.57)} & \textbf{73.81 (+0.33)} & \textbf{76.54 (+1.03)} & \textbf{76.54 (+1.06)} & \textbf{74.62 (+0.48)} & \textbf{74.37 (+0.43)}  \\ 
\cmidrule{2-8}
                         & AT~\cite{zagoruyko2016paying}       & 70.55~                 & 72.31~                 & 73.44~                 & 74.08~                 & 72.77~                 & 71.43~                  \\
                         & +DFKD    & \textbf{70.98 (+0.43)} & \textbf{72.93 (+0.62)} & \textbf{74.31 (+0.87)} & \textbf{75.15 (+1.07)} & \textbf{73.09 (+0.32)} & \textbf{71.64 (+0.21)}  \\ 
\cmidrule{2-8}
                         & ReviewKD~\cite{chen2021distilling} & 71.89~                 & 73.89~                 & 75.63~                 & 76.12~                 & 75.09~                 & 74.84~                  \\
                         & +DFKD    & \textbf{72.61 (+0.72)} & \textbf{74.63 (+0.74)} & \textbf{77.41 (+1.78)} & \textbf{77.02 (+0.90)} & \textbf{75.63 (+0.54)} & \textbf{75.31 (+0.47)} \\
\bottomrule
\end{tabular}
\end{table*}

%% file: table/heme_cifar.tex
\begin{table*}
\centering
\caption{Results of top-1 accuracy (\%) for heterogeneous architectures on CIFAR-100.}
\label{heme_cifar}
\begin{tabular}{ll|ccccc} 
\toprule
                          & Teacher  & ResNet$32\times4$      & WRN-40-2               & VGG13                  & ResNet50               & ResNet$32\times4$       \\
                          &          & 79.42                  & 75.61                  & 74.64                  & 79.34                  & 79.42                   \\
                          & Student  & ShuffleNetV1           & ShuffleNetV1           & MobileNetV2            & MobileNetV2            & ShuffleNetV2            \\
                          &          & 70.50~                 & 70.50~                 & 64.60~                 & 64.60~                 & 71.82~                  \\ 
\cmidrule{1-2}\cmidrule{3-7}
\multirow{3}{*}{Logits}   & KD~\cite{hinton2015distilling}       & 74.07                  & 74.83                  & 67.37                  & 67.35                  & 74.45                   \\
                          & DKD~\cite{zhao2022decoupled}      & 76.45                  & 76.70                  & 69.71                  & 70.35                  & 77.07                   \\
                          & MLKD~\cite{jin2023multi}     & 77.18                  & 77.44                  & 70.57                  & 71.04                  & 78.44                   \\
                           & ND$^*$~\cite{Sun2024Logit}     & 77.01                  & 77.25                  & 70.94                  & 71.19                  & 78.76  \\
\bottomrule
\multirow{8}{*}{Features} & FitNet~\cite{romero2014fitnets}   & 73.59~                 & 73.73~                 & 64.14~                 & 63.16~                 & 73.54~                  \\
                          & +DFKD    & \textbf{74.93 (+1.34)} & \textbf{75.65 (+1.92)} & \textbf{66.32 (+2.18)} & \textbf{67.12 (+3.96)} & \textbf{74.21 (+0.67)}  \\ 

\cmidrule{2-7}
                          & AT~\cite{zagoruyko2016paying}       & 71.73~                 & 73.32~                 & 69.40~                 & 68.58~                 & 72.73~                  \\
                          & +DFKD    & \textbf{73.24 (+1.51)} & \textbf{75.09 (+1.77)} & \textbf{69.64 (+0.24)} & \textbf{68.75 (+0.17)} & \textbf{73.43 (+0.70)}  \\ 
\cmidrule{2-7}
                          & ReviewKD~\cite{chen2021distilling} & 77.45~                 & 77.14~                 & 70.37~                 & 69.89~                 & 77.78~                  \\
                          & +DFKD    & \textbf{78.12 (+0.67)} & \textbf{77.32 (+0.18)} & \textbf{70.79 (+0.42)} & \textbf{71.10 (+1.21)} & \textbf{78.64 (+0.86)}  \\
\cmidrule{2-7}
                          & CRD~\cite{tian2019contrastive}      & 75.11~                 & 76.05~                 & 69.73~                 & 69.11~                 & 75.65~                  \\
                          & +DFKD    & \textbf{75.73 (+0.62)} & \textbf{77.29 (+1.24)} & \textbf{70.44 (+0.71)} & \textbf{69.57 (+0.46)} & \textbf{76.24 (+0.59)}  \\ 
\bottomrule
\end{tabular}
\end{table*}

%% file: table/imagenet.tex
\begin{table*}[tp]
\centering
\caption{Top-1 and Top-5 accuracy (\%) of student networks on ImageNet validation set.}\label{imagenet}
\begin{tabular}{ccc|cccccc} 
\toprule
\multicolumn{1}{l}{} & Teacher & Student              & AT     & +DFKD & CRD    & +DFKD & ReviewKD & +DFKD                       \\ 
\midrule
\multicolumn{1}{l}{} &         & \multicolumn{1}{c}{} & \multicolumn{6}{c}{ResNet34 as the teacher, ResNet18 as the student}     \\ 
\midrule
top-1 & 73.31~  & 69.75~  
& 70.69~ & \textbf{72.13 \textcolor{black}{(+1.44)}}  & 71.17~ & \textbf{71.86 \textcolor{black}{(+0.69)}}  & 71.61~ & \textbf{72.53 \textcolor{black}{(+0.92)}} \\
top-5 & 91.42~  & 89.07~  & 90.01~ & \textbf{91.34 \textcolor{black}{(+1.33)}}  & 90.13~ & \textbf{90.71 \textcolor{black}{(+0.58)}}  & 90.51~ & \textbf{91.32 \textcolor{black}{(+0.81)}}                            \\ 
\cline{1-9}
\\[-1em] 
\cline{1-9}
\multicolumn{1}{c}{} &         & \multicolumn{1}{c}{} & \multicolumn{6}{c}{ResNet50 as the teacher, MobileNetV1 as the student}  \\ 
\midrule
top-1  & 76.16~  & 68.87~  & 69.56~ &  \textbf{69.83 \textcolor{black}{(+0.27)}}     & 71.37~ & \textbf{71.60 \textcolor{black}{(+0.23)}}      & 72.56~   & \textbf{73.02 \textcolor{black}{(+0.48)}}                            \\
top-5  & 92.86~  & 88.76~  & 89.33~ &  \textbf{89.58 \textcolor{black}{(+0.25)}}     & 90.41~ & \textbf{90.69 \textcolor{black}{(+0.28)}}      & 91.00~   & \textbf{91.22 \textcolor{black}{(+0.22)}}                        \\
\bottomrule
\end{tabular}
\end{table*}

%% file: table/coco.tex
\begin{table*}
\centering
\caption{Object detection results on MS-COCO. We take Faster-RCNN with FPN as the backbones.
}
\label{COCO}
\begin{tabular}{l|cc|cc|cc} 
\toprule
         &AP   &AP50    &AP &AP50   &AP &AP50                        \\ 
\midrule
         & \multicolumn{2}{c|}{ResNet101 \& ResNet18} & \multicolumn{2}{c|}{ResNet101 \& ResNet50} & \multicolumn{2}{c}{ResNet50 \& MobileNetV2}  \\ 
\midrule
Teacher  &42.04        & 62.48                    & 42.04        & 62.48                    & 40.22        & 61.02                       \\
Student  &33.26        & 53.61                    & 37.93        & 58.84                    & 29.47        & 48.87                       \\ 
\midrule
FitNet   &34.13        & 54.16                    & 38.76        & 59.62                    & 30.2         & 49.8                        \\
+DFKD         & \textbf{35.09 \textcolor{black}{(+0.96)}} & \textbf{54.93 \textcolor{black}{(+0.77)}}             & \textbf{39.43 \textcolor{black}{(+0.67)}} & \textbf{60.08 \textcolor{black}{(+0.46)}}             & \textbf{31.43 \textcolor{black}{(+1.23)}} & \textbf{50.85 \textcolor{black}{(+1.05)}}                \\ 
\midrule
FGFI     & 35.44        & 55.51                    & 39.44        & 60.27                    & 31.16        & 50.68                       \\
+DFKD         & \textbf{36.32 \textcolor{black}{(+0.88)}}  & \textbf{56.32 \textcolor{black}{(+0.81)}}             & \textbf{39.78 \textcolor{black}{(+0.34)}} & \textbf{60.64 \textcolor{black}{(+0.37)}}             & \textbf{32.02 \textcolor{black}{(+0.86)}} & \textbf{51.23 \textcolor{black}{(+0.55)}}                \\ 
\midrule
ReviewKD & 36.75        & 56.72                    & 40.36        & 60.97                    & 33.71        & 53.15                       \\
+DFKD         & \textbf{37.21 \textcolor{black}{(+0.46)}} & \textbf{57.52 \textcolor{black}{(+0.80)}}             & \textbf{40.43 \textcolor{black}{(+0.07)}} & \textbf{61.86 \textcolor{black}{(+0.89)}}             & \textbf{34.24 \textcolor{black}{(+0.53)}} & \textbf{54.29 \textcolor{black}{(+1.14)}}                \\
\bottomrule
\end{tabular}
\end{table*}

%% file: table/ablation_study.tex
\begin{table}
\centering
\caption{Ablation for different modules in DFKD. 
R-324 and R-84 respectively denote ResNet $32\times4$ and ResNet $8\times4$.}\label{main_ablation}
\begin{tabular}{c|cccc} 
\toprule
\multicolumn{1}{l|}{Teacher \& Student} &DAFA &$\mathcal{L}_{\text{con}}$ &DAM &Accuracy (\%)                 \\ 
\midrule
\multirow{4}{*}{R-324 \& R-84}             &-            &-              &-            & \multicolumn{1}{c}{75.63}  \\
                                      &\ding{51}    & \ding{55}     & \ding{55}    &   76.85                   \\
                                      &\ding{51}    & \ding{51}     & \ding{55}    &   77.01                   \\
                                      & \ding{51}   & \ding{51}     & \ding{51}    &   \textbf{77.41}                   \\ 
\midrule
\multirow{4}{*}{VGG13 \& MV2}          &-              & -           & -    & \multicolumn{1}{c}{70.37}  \\
                                      & \ding{51}     & \ding{55}   & \ding{55}    &   70.52                   \\
                                      & \ding{51}     & \ding{51}   & \ding{55}    &   70.67                   \\
                                      & \ding{51}     & \ding{51}   & \ding{51}    &   \textbf{70.79}                   \\
\bottomrule
\end{tabular}
\end{table}

%% file: table/search_epochs.tex
\begin{table}
\centering
\caption{Influence of search epoch number in DAFA.}
\label{search_epoch}
\begin{tabular}{c|cc} 
\hline
Search Epochs & R110 \&  R32 & VGG13 \& MV2  \\ 
\hline
10            & 74.21                   & 70.43               \\
20            & \textbf{74.63}          & \textbf{70.79}      \\
25            & 74.49                   & 70.71               \\
30            & 74.52                   & 70.59               \\
\hline
\end{tabular}
\end{table}

%% file: table/diff_aug.tex
\begin{table}
\centering
\caption{Experimental comparison results of different types of augmentations based on FitNet for various teacher-student pairs.
}
\label{dif_aug}
\begin{tabular}{l|ccc} 
\toprule
Method      & \begin{tabular}[c]{@{}c@{}}ResNet56 \\ResNet20\end{tabular} & \begin{tabular}[c]{@{}c@{}}ResNet32x4 \\ResNet8x4\end{tabular} & \begin{tabular}[c]{@{}c@{}}WRN-40-2\\ShuffleNetV1\end{tabular}  \\ 
\hline
Data Aug    & 67.50                                                       & 72.25                                                       & 69.27   \\
Feature Aug & 69.72                                                       & 73.28                                                       & 72.52   \\
Aug Search (Ours)  & \textbf{71.54}                                              & \textbf{74.41}                                              & \textbf{75.65}   \\
\bottomrule
\end{tabular}
\end{table}